\documentclass[a4paper,twocolumn]{esapub2005}
\usepackage{times}
\usepackage{natbib}
\usepackage{graphicx}
\usepackage{amsmath,amssymb,amsfonts}
\usepackage{hyperref}
\usepackage{xcolor}
\usepackage{subcaption}
\usepackage{booktabs}
\usepackage{multirow}
\usepackage{enumitem}

\title{Learning Tool-Aware Adaptive Compliant Control\\ for Autonomous Regolith Excavation}
\author{Andrej Orsula\textsuperscript{\normalfont{1}}}
\author{Matthieu Geist\textsuperscript{\normalfont{2}}}
\author{Miguel Olivares-Mendez\textsuperscript{\normalfont{1}}}
\author{Carol Martinez\textsuperscript{\normalfont{1}}}
\affil{\textsuperscript{\normalfont{1}}University of Luxembourg\\\textsuperscript{\normalfont{2}}Earth Species Project}

\begin{document}
\hypersetup{
    pdfinfo ={
            Title={Learning Tool-Aware Adaptive Compliant Control for Autonomous Regolith Excavation},
            Author={Andrej Orsula, Matthieu Geist, Miguel Olivares-Mendez, Carol Martinez},
            Keywords={Reinforcement Learning, Compliant Control, Resource Extraction},
            Creator={LaTeX},
        }
}


\maketitle

\begin{abstract}
    Autonomous regolith excavation is a cornerstone of in-situ resource utilization for a sustained human presence beyond Earth. However, this task is fundamentally hindered by the complex interaction dynamics of granular media and the operational need for robots to use diverse tools. To address these challenges, this work introduces a framework where a model-based reinforcement learning agent learns within a parallelized simulation. This environment leverages high-fidelity particle physics and procedural generation to create a vast distribution of both lunar terrains and excavation tool geometries. To master this diversity, the agent learns an adaptive interaction strategy by dynamically modulating its own stiffness and damping at each control step through operational space control. Our experiments demonstrate that training with a procedural distribution of tools is critical for generalization and enables the development of sophisticated tool-aware behavior. Furthermore, we show that augmenting the agent with visual feedback significantly improves task success. These results represent a validated methodology for developing the robust and versatile autonomous systems required for the foundational tasks of future space missions.
    \textit{The source code is available at~\href{https://github.com/AndrejOrsula/space_robotics_bench}{github.com/AndrejOrsula/space\_robotics\_bench}.}
\end{abstract}

\section{Introduction}

\begin{figure}[t]
    \centering
    \includegraphics[width=\linewidth]{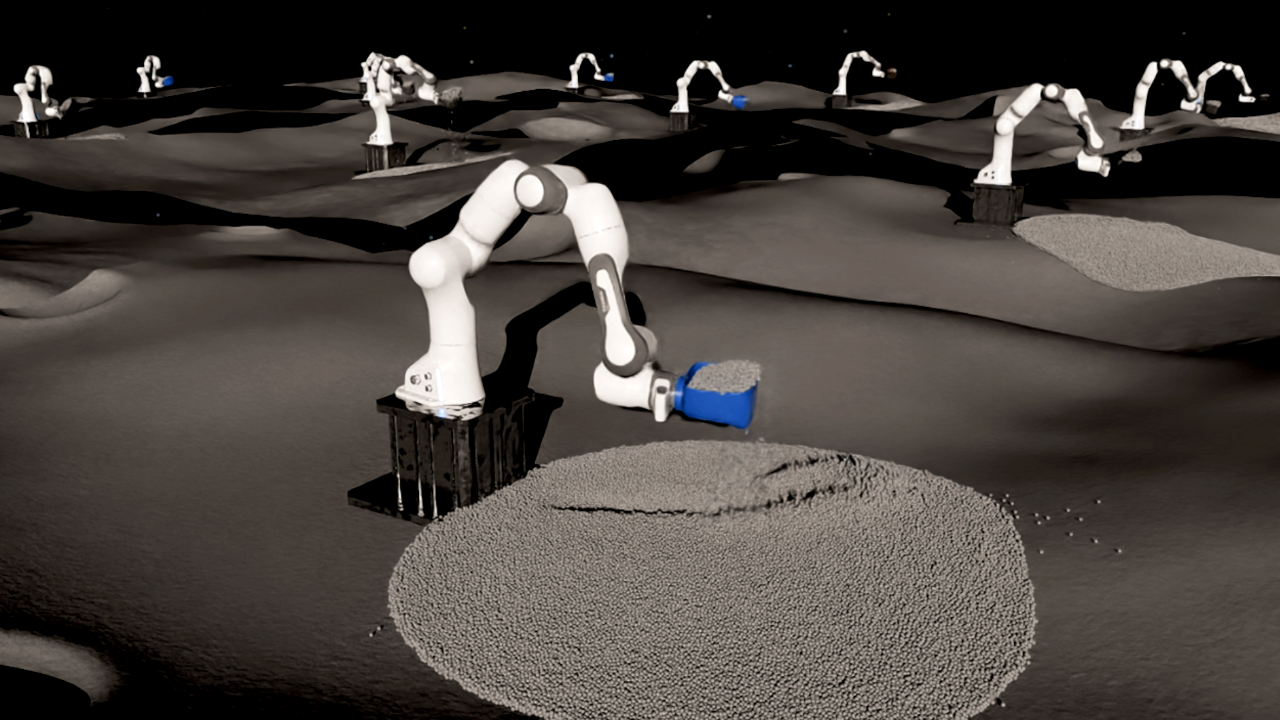}
    \caption{A generalizable excavation skill is learned within the Space Robotics Bench by training agents on a vast distribution of randomized scenarios with procedurally generated tools, which fosters an adaptive compliance policy for successful tool-aware interaction.}
    \label{fig:concept}
    \vspace{-0.75\baselineskip}
\end{figure}

The success of ambitious international efforts like the Artemis program depends on establishing a sustained human and robotic presence on the Moon~\cite{nasa2020artemis}. In-Situ Resource Utilization (ISRU) is a foundational principle for this endeavor~\cite{crawford2015lunar}. The ability to harness local materials is a critical enabler for long-term missions because it significantly reduces reliance on complex and costly Earth-based supply chains. At the forefront of ISRU is the autonomous excavation of lunar regolith. This single capability underpins a wide range of future operations, including the large-scale construction of infrastructure and the extraction of vital resources like water, ice, and oxygen~\cite{zhang2019progress}. The successful execution of these demanding tasks over multi-year missions will depend on the development of robotic systems that can operate reliably and adaptively with minimal human supervision.

Developing an autonomous excavation system presents a formidable set of interconnected challenges. The first major obstacle is the complex and unpredictable physics of lunar environments covered with regolith, which is a fine-grained granular material whose mechanical behavior defies simple analytical models~\cite{mitchell1972mechanical,slyuta2014physical}. The properties of this material are known to be highly variable across the lunar surface, yet our current understanding is derived from a small number of past missions~\cite{heiken1991lunar}. Future operations will target unexplored regions where key characteristics like particle size, compaction, and the presence of ice are largely unknown. This fundamental uncertainty renders it impossible to create a single, accurate, mission-specific model, as a robotic manipulator digging into this medium will encounter highly variable and often large resistance forces~\cite{connolly2023engineering}. Traditional rigid control strategies are inherently brittle under such unpredictable conditions and can easily lead to mission-critical failures. Such failures include the manipulator stalling, drawing excessive power, or sustaining catastrophic damage to its hardware. A robust system must therefore be able to manage these unmodeled physical interactions through compliant behavior rather than rigid control~\cite{hogan1985impedance}.

The need for operational versatility compounds this physical challenge. A single robotic platform will be expected to perform a variety of excavation tasks over its operational lifetime. These diverse tasks will naturally require different tools, such as wide scoops for bulk material transfer or specialized toothed buckets for breaking compacted ground. This need for adaptation extends beyond intentional tool changes to include unintentional ones as well. The abrasive nature of regolith can cause significant wear and tear over a long-duration mission~\cite{gaier2007effects}, leading to unforeseen changes in tool geometry, such as blunted edges, broken teeth, or even partial loss of the tool. A practical autonomous system cannot be pre-programmed for a fixed set of tool configurations. It must possess a general excavation skill that allows it to intelligently adapt its strategy to its changing physical form, whether from a deliberate tool swap or gradual degradation. The harsh lunar environment further constrains these operations. The goal of maximizing collected regolith must be carefully balanced against the critical need to minimize abrasive dust generation, which poses a significant and persistent threat to nearby actuators and sensors.

This work introduces a learning-based framework where an autonomous agent acquires generalizable excavation skills to address these challenges. Our approach, conceptually illustrated in Figure~\ref{fig:concept}, centers on training the agent in a high-fidelity simulation featuring a rich distribution of procedurally generated scenarios. The agent learns to master complex physical interactions not through rigid commands but by dynamically modulating its own physical compliance. This method enables the development of sophisticated, tool-aware strategies that are robust and efficient. This paper makes several key contributions. We first present a comprehensive simulation framework for learning regolith excavation within the Space Robotics Bench (SRB)~\cite{orsula2024towards}. This environment integrates procedural generation of both lunar terrains and diverse excavation tools inside a high-fidelity particle physics simulation. We then demonstrate that training an agent with a procedural distribution of tools is critical for achieving a robust and generalizable policy. We also provide an analysis of the emergent behaviors that result from learned adaptive compliance. Finally, this investigation explores how augmenting the agent with visual feedback from depth perception enhances its ability to coordinate the complex interactions between tools and regolith.

\begin{figure*}[t]
    \centering
    \subcaptionbox{Parametric pipelines generate diverse lunar terrains and excavation tools, creating a near-infinite source of unique training assets.\label{fig:procgen_explained}}{\includegraphics[width=0.4935\linewidth]{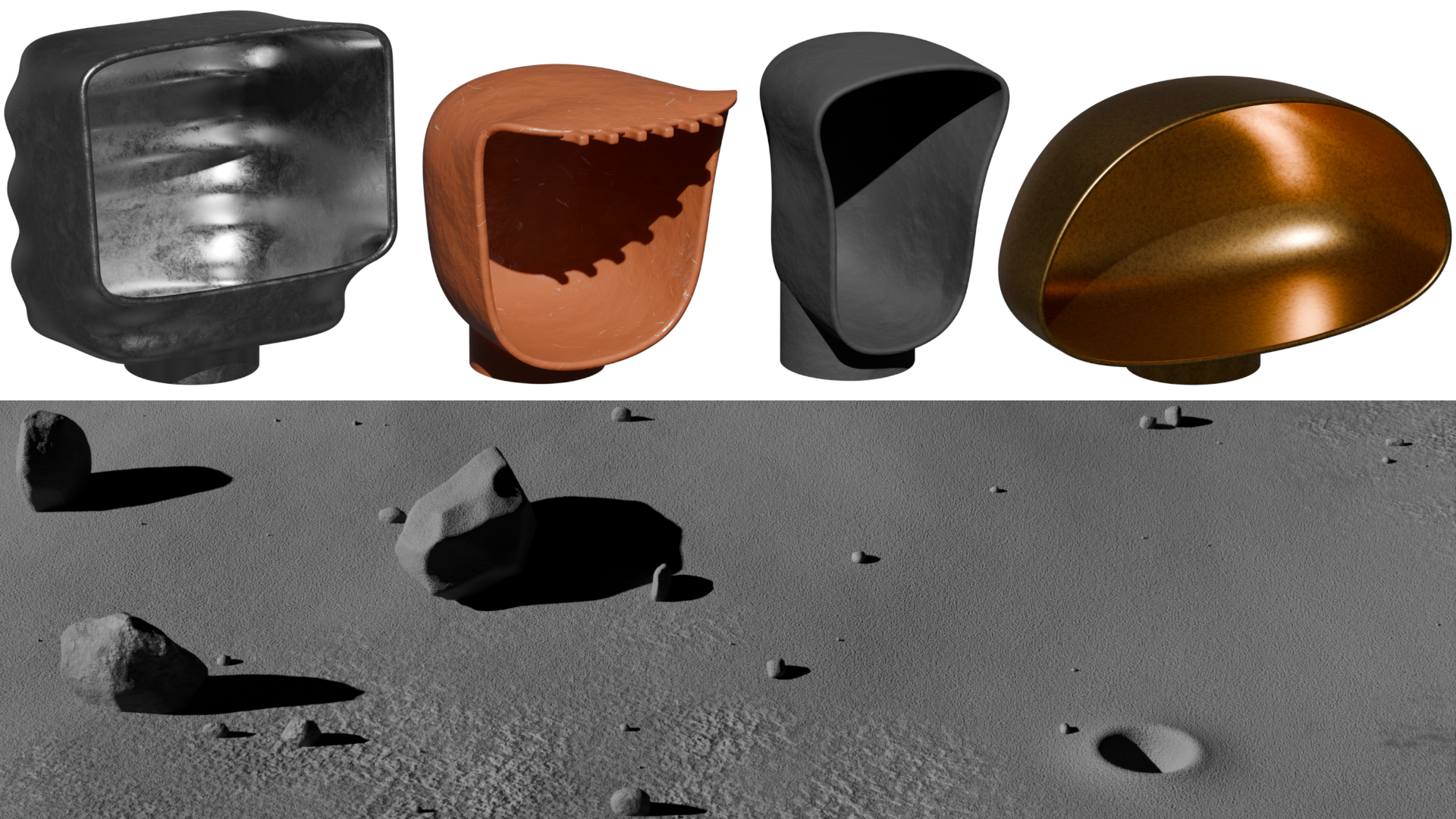}}
    \hfill
    \subcaptionbox{A visualization of 16 parallel training environments with procedural diversity that creates a massively diverse learning experience.\label{fig:domain_rand_explained}}{\includegraphics[width=0.4935\linewidth]{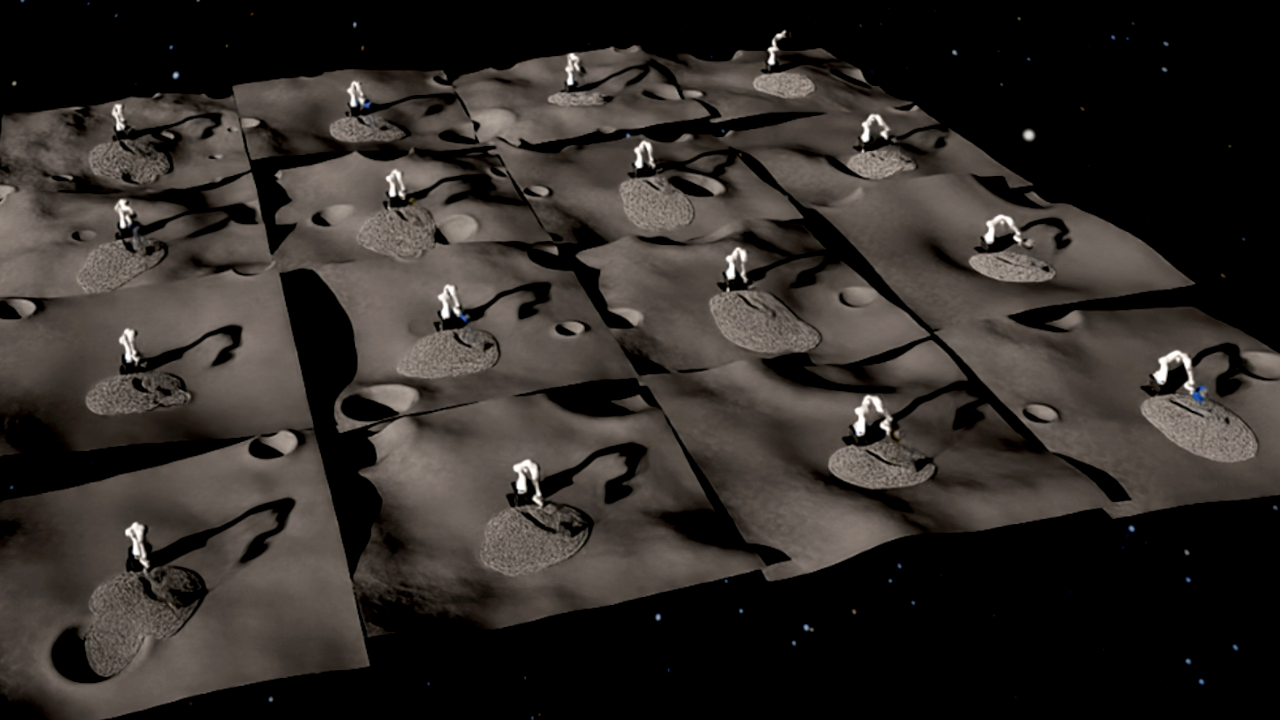}}
    \caption{Our procedural paradigm, with the SimForge engine populating many parallel environments in the Space Robotics Bench to facilitate the rich and varied training experience that is essential for learning truly generalizable skills.}
    \label{fig:procedural_testbed}
    \vspace{-0.75\baselineskip}
\end{figure*}

\section{Related Work}

The field of terramechanics traditionally addresses the challenge of robotic interaction with granular media~\cite{ishigami2007terramechanics}. It provides analytical models that can predict interaction forces and offer valuable insights for tasks like rover traversal. However, these models are often insufficient for the high-dimensional dynamics of manipulation~\cite{sotiropoulos2021methods}. The reaction forces during an excavation task depend on specific tool geometry, the precise trajectory through the medium, and highly variable local regolith properties. The behavior of granular material is highly non-linear, and its properties are difficult to model, especially when combined with the uncertain conditions of space. Creating accurate analytical models for such intricate interactions is therefore exceptionally difficult. This difficulty motivates our data-driven approach, where a control policy can implicitly learn the nuances of excavation through direct experience.

Due to the limited availability of data, the development of such data-driven policies for space robotics is only feasible through high-fidelity simulation. This work is a direct application of the principles behind SRB, our simulation framework designed to facilitate robot learning for diverse space applications~\cite{orsula2024towards}. A core philosophy in modern robotics is that policy robustness stems from a massive diversity of experience rather than a single perfect digital twin. Foundational techniques like domain randomization have proven effective in bridging the sim-to-real gap by varying simulation parameters to enforce robustness to uncertainty~\cite{tobin2017domain}. We extend this principle with procedural content generation (PCG), which allows us to create a nearly infinite distribution of unique training scenarios. While the majority of prior research has applied PCG to simple 2D environments~\cite{cobbe2020leveraging,koutras2021marsexplorer}, we have demonstrated the effectiveness of this approach for contact-rich assembly~\cite{orsula2024leveraging} and sim-to-real transfer of a navigation policy~\cite{orsula2025sim2dust}. The key distinction in this work is that we apply PCG directly to the robot's own physical embodiment. We procedurally generate a wide variety of excavation tool geometries, rather than merely creating diverse environments. This encourages the agent to learn a generalizable skill not tied to a specific end-effector.

Reinforcement Learning (RL) has emerged as a powerful paradigm for discovering sophisticated strategies in problems where optimal behaviors are non-intuitive~\cite{sutton2018reinforcement}. Its potential for autonomous excavation has been demonstrated in terrestrial applications, such as controlling wheel loaders to perform soil scooping~\cite{azulay2021wheel} and planning excavation trajectories for rigid objects~\cite{lu2022excavation}. The complexity of interaction dynamics has also been explored in related domains such as water scooping~\cite{niu2023goats}, where curriculum learning was employed to accelerate the learning process. Our methodology is distinct from prior work in two critical ways. First, we employ a model-based RL algorithm. Methods like DreamerV3 learn a predictive world model that can significantly improve sample efficiency, allowing an agent to plan more effectively in environments with complex physical dynamics~\cite{hafner2025mastering}. Second, we focus on the control representation that defines how the agent physically executes its decisions. Rigid kinematic control is often brittle in contact-rich scenarios~\cite{hogan1985impedance}. To address this, Operational Space Control (OSC) provides a principled framework for managing interaction forces through direct control of end-effector compliance~\cite{khatib1987unified}. Beyond using manually tuned gains, OSC can be extended with learning-based approaches that dynamically adjust stiffness and damping based on sensory feedback~\cite{peters2006learning}. This adaptability is key to enhancing the performance of autonomous systems in dynamic tasks. While prior research has combined OSC with model-free RL for terrestrial tasks~\cite{wong2022oscar}, its application to space robotics remains unexplored. The central innovation of our work is the fusion of these two concepts. We leverage a model-based RL agent to dynamically learn and modulate the stiffness and damping parameters of its own OSC controller. This creates a tight synergy between high-level learning and low-level physical control, enabling the agent to discover adaptive compliance strategies tailored to the unique challenges of robotic excavation in space.

\section{Methodology}

Our approach is founded on an integrated framework that combines a scalable and diverse simulation testbed with a sophisticated learning methodology. The testbed is designed to generate the rich experience necessary for training a robust policy, while the learning methodology then enables an agent to translate that experience into a generalizable and physically adaptive excavation skill.

\subsection{Procedural Testbed for Regolith Excavation}

All agent training and evaluation occur within the Space Robotics Bench~\cite{orsula2024towards}. This simulation framework for robot learning in space is built on top of NVIDIA Isaac Lab~\cite{mittal2023orbit}, which provides a scalable foundation through its massively parallel GPU-accelerated physics and rendering. This high-throughput architecture is essential for our procedural workflow to facilitate the generation of diverse experiences at scale, which is crucial for generating the vast and varied datasets required by modern RL algorithms. To accurately capture the complex physics of excavation, we model the regolith as a granular medium using high-fidelity particle physics that is based on extended position-based dynamics (XPBD)~\cite{macklin2019small}. This physics-based approach is crucial for representing the complex, non-linear dynamics of the material. It also provides the realistic and highly variable force feedback that the manipulator experiences during interaction, which is a necessary component for learning a physically grounded skill.

The foundation of our approach for developing a robust policy is training with diverse data. We achieve this through two synergistic mechanisms that are depicted in Figure~\ref{fig:procedural_testbed}. First, we generate unique structural assets for each parallel training instance through our integrated procedural engine SimForge~\cite{orsula2025advancing}. The procedural pipelines for both lunar terrains and excavation tools are hand-crafted using the modular Geometry Nodes system within Blender. This node-based system allows for the creation of complex parametric rules that deterministically define an entire family of 3D models. This geometric generation is combined with procedural materials based on Shader Nodes. These materials are then baked into a set of physically-based rendering (PBR) textures before the final asset is spawned within the simulation. To produce a near-infinite variety of unique assets, the engine automatically samples the pipeline parameters from a seeded random distribution for each exported variant. The resulting assets are showcased in Figure~\ref{fig:procgen_explained}.

\begin{figure}[ht]
    \centering
    \includegraphics[width=\linewidth]{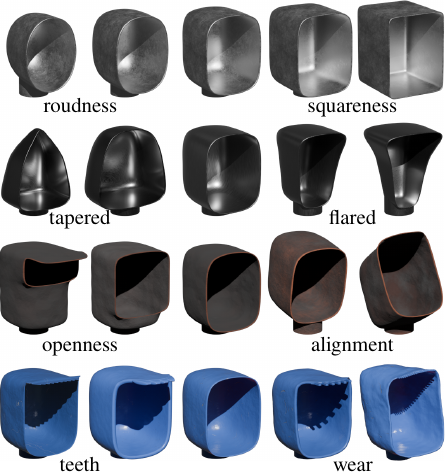}
    \caption{A grid of procedurally generated excavation tool geometries. Each row isolates the effect of modulating selected parameters. This demonstrates the fine-grained control over the resulting tool morphology.}
    \label{fig:tool_variety}
    \vspace{-0.75\baselineskip}
\end{figure}

The procedural generation of the excavation tools themselves is a central innovation of our methodology. The tool geometry pipeline features over forty parameters that control a wide array of morphological features. These parameters modulate the fundamental size and shape of the scoop, including its width, depth, and curvature. They also control features like mounting interface alignment, the number and shape of teeth, and even the overall deterioration of the tool to simulate wear from long-term use. As showcased in Figure~\ref{fig:tool_variety}, this level of control enables us to generate a continuous spectrum of end-effectors. These tools range from wide scoops for bulk material transfer to narrow buckets with sharp serrated edges for breaking compacted regolith with elevated cohesion. By training on this procedural distribution of tools, the agent is forced to learn an abstract and generalizable understanding of excavation. It cannot rely on a single memorized trajectory. It must instead develop an adaptive strategy that can be generalized to the specific geometry of the tool it currently holds.

A complementary layer of diversity is added through extensive domain randomization of the simulation's physical and visual parameters. Physical randomization targets the underlying dynamics to build robustness against unmodeled physical effects and environmental uncertainty. This includes randomizing factors such as the density, friction, and cohesion of individual regolith particles. The magnitude of the gravity is also sampled from a distribution and varied at random intervals to simulate operation in different lunar regions. Visual randomization is applied to the ambient illumination to match the harsh lighting of the lunar day.
The true power of our methodology is realized through its scalability. Each of the parallel environments in Figure~\ref{fig:domain_rand_explained} is not just structurally unique due to PCG, but it is also parametrically unique in every episode due to domain randomization. This comprehensive combination of procedural and parametric diversity creates the rich, high-variance distribution of training scenarios required to forge a truly generalizable policy.

\subsection{Learning Tool-Aware Adaptive Control}\label{sec:learning_methodology}

We formulate the excavation task as a Partially Observable Markov Decision Process (POMDP) because the agent cannot access the complete state of the world. Key physical properties, such as the shape of the terrain and the position of every regolith particle, are not directly observable. Similarly, we do not provide the agent with the intrinsic properties of local regolith, such as friction and cohesion, due to the difficulty of measuring these properties in situ. The unobservability is compounded by the fact that the agent does not have access to the geometry of its equipped tool. This creates a challenging setting where the agent must infer hidden states from its history of physical interactions.

We design the task as a finite-horizon problem. A single episode requires the agent to learn a sequence of actions that successfully lifts and stabilizes a significant volume of regolith. Each episode is truncated after a maximum duration of 15 seconds, and we use no other state-based termination conditions. This fixed-duration formulation is a deliberate design choice. It ensures the learned policy represents a modular, reusable skill. The resulting policy can then be integrated as a low-level primitive within a larger hierarchical control system, which could execute long-horizon missions by repeatedly invoking this foundational skill.

The objective is to learn a policy that maps a sequence of observations to actions that maximize the expected cumulative reward. We employ the DreamerV3~\cite{hafner2025mastering} model-based RL algorithm to learn the control policy. Its ability to learn a predictive world model is exceptionally well-suited for mastering the complex dynamics of granular media~\cite{orsula2025sim2dust}. Its recurrent world model architecture serves as a memory that allows the agent to integrate information over time and infer the hidden states of the environment. Furthermore, its high sample efficiency is crucial for training with the computationally intensive simulation of granular media.

The agent's observation space is based on proprioception, which includes the manipulator's normalized joint positions and joint torque measurements. The end-effector pose from Forward Kinematics is also provided, with orientation encoded using a 6D representation~\cite{zhou2019continuity}. To investigate the impact of augmenting the agent with sensory feedback, we also evaluate a variant that incorporates $128 \times 128$~px depth maps of its immediate workspace from two perspectives. We deliberately omit any privileged information regarding the tool geometry. This decision reflects a more realistic scenario and prevents the agent from overfitting to our specific procedural pipeline. Instead, the agent must infer the tool's geometry through physical interactions, which is a more challenging but ultimately more generalizable approach.

\begin{figure}[ht]
    \centering
    \includegraphics[width=\linewidth]{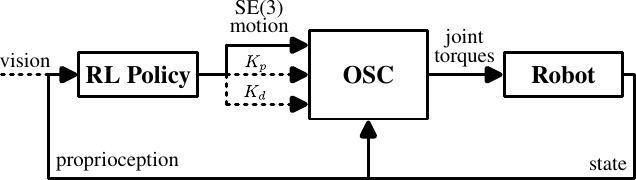}
    \caption{The action space of our agent combines SE(3) motion commands with stiffness and damping gains to achieve learned adaptive compliance through OSC.}
    \label{fig:action_space}
    \vspace{-0.75\baselineskip}
\end{figure}

The actions of the agent directly integrate with OSC, as illustrated in Figure~\ref{fig:action_space}. The core concept involves using OSC to provide software-defined compliance through adjustable stiffness and damping gains, which can be dynamically modulated by the agent at every time step based on task requirements. In our experiments, we investigate three distinct control strategies that primarily differ in how compliance is handled and learned.
\begin{description}[nosep,topsep=-0.5em,leftmargin=1em]
    \item \textbf{\textsc{ik}} [$\text{dim}(\mathcal{A})\hspace{-0.5mm}=\hspace{-0.5mm}6$]: A non-compliant baseline using standard differential Inverse Kinematics.
    \item \textbf{\textsc{osc-part}} [$\text{dim}(\mathcal{A})\hspace{-0.5mm}=\hspace{-0.5mm}12$]: OSC agent that learns to modulate only the stiffness ($K_p$) gains.
    \item \textbf{\textsc{adaptive}} [$\text{dim}(\mathcal{A})\hspace{-0.5mm}=\hspace{-0.5mm}18$]: OSC agent that learns to modulate both stiffness ($K_p$) and damping ($K_d$) gains.
\end{description}

Across all strategies, a core component of the action space $\mathcal{A}$ includes the desired end-effector translational and rotational displacements in SE(3). For the adaptive strategies, the action space is augmented accordingly, where \textsc{osc-part} adds six dimensions for the stiffness gains $K_p$, while \textsc{adaptive} adds another six dimensions for the damping gains $K_d$. This allows the agent to actively control the robot's impedance based on its learned policy. For instance, the agent can choose to make the end-effector rigid when moving through free space and then switch to a more compliant mode when interacting with the regolith. This dynamic modulation of compliance is a key innovation that enables the agent to adapt its physical interaction style to the specific tool geometry and the current state of the excavation task.

We employ a composite reward function to guide the agent towards the desired excavation strategy. The total reward provided at each time step is defined as a sum of several components.
\setlength\abovedisplayskip{0.5em}
\setlength\belowdisplayskip{0.5em}
\begin{equation*}
    R_\text{total} = R_\text{approach} + R_\text{lift} + R_\text{stabilize} - P_\text{create\_dust} - P_\text{jerk}
\end{equation*}
The primary objective is captured by the two positive terms $R_\text{lift}$ and $R_\text{stabilize}$. $R_\text{lift}$ provides a reward proportional to the number of regolith particles successfully lifted to a target height. $R_\text{stabilize}$ then provides a substantial bonus for achieving a stable lift, specifically rewarding particles that are within the target height range and have near-zero velocity. This structure heavily incentivizes a smooth, controlled final state where the excavated material is held stationary. Early-stage learning is encouraged by the $R_\text{approach}$ term, which facilitates approaching the regolith pile and maintaining a correct top-down scoop orientation. These objectives are balanced by the penalty term $P_\text{create\_dust}$, which penalizes dust creation proportional to regolith particle velocity. Lastly, the term $P_\text{jerk}$ encourages smoother motion by penalizing action rate and excessive joint torques. This comprehensive reward structure provides a dense learning signal that guides the agent toward mastering the nuanced skill of autonomous excavation. The particle-related reward components are computationally expensive, as the data transfer alone for reading the positions and velocities of particles would require a minimum bandwidth of $1.2$~GB/s for every $1$M particles. Therefore, these terms are computed only once every 2 seconds and cached for intermediate steps. This design improves simulation performance but introduces reward delays, which makes the notion of credit assignment more challenging.

\begin{table*}[t]
    \centering
    \caption{Zero-shot generalization performance of all evaluated agents on the held-out set of $8$ novel tool geometries.}
    \label{tab:main_results}
    \begin{tabular}{
            @{}
            r@{\ }l
            c
            c
            c
            @{}
        }
        \toprule

        \multicolumn{2}{@{}c}{\multirow{2}{*}{\textbf{Agent Configuration}}} & \multicolumn{1}{c}{\textbf{Excavated Volume}} & \multicolumn{1}{c}{\textbf{Dust Generated}}            & \multicolumn{1}{c@{}}{\textbf{Motion Jerk}}                                                                     \\
                                                                             &                                               & \multicolumn{1}{c}{\color{gray}[unit: L]}              & \multicolumn{1}{c}{\color{gray}(normalized)}           & \multicolumn{1}{c@{}}{\color{gray}(normalized)}        \\
        \midrule
        \textsc{ik} \vline                                                   & IK\hspace{20.95mm}+ PCG Tools                 & $0.13 \hspace{-0.5mm}\pm\hspace{-0.5mm} 0.03$          & $1.00 \hspace{-0.5mm}\pm\hspace{-0.5mm} 0.03$          & $1.00 \hspace{-0.5mm}\pm\hspace{-0.5mm} 0.68$          \\
        \textsc{osc-part} \vline                                             & OSC ($K_p$)\hspace{9.99mm}+ PCG Tools         & $0.09 \hspace{-0.5mm}\pm\hspace{-0.5mm} 0.06$          & $0.97 \hspace{-0.5mm}\pm\hspace{-0.5mm} 0.04$          & $0.41 \hspace{-0.5mm}\pm\hspace{-0.5mm} 0.27$          \\
        \textsc{specialist} \vline                                           & OSC ($K_p$ \& $K_d$) + Static Tool            & $0.02 \hspace{-0.5mm}\pm\hspace{-0.5mm} 0.04$          & $0.91 \hspace{-0.5mm}\pm\hspace{-0.5mm} 0.03$          & $0.39 \hspace{-0.5mm}\pm\hspace{-0.5mm} 0.18$          \\
        \textsc{adaptive} \vline                                             & OSC ($K_p$ \& $K_d$) + PCG Tools              & $0.14 \hspace{-0.5mm}\pm\hspace{-0.5mm} 0.07$          & $0.94 \hspace{-0.5mm}\pm\hspace{-0.5mm} 0.04$          & $0.44 \hspace{-0.5mm}\pm\hspace{-0.5mm} 0.17$          \\
        \textsc{visual} \vline                                               & OSC ($K_p$ \& $K_d$) + PCG Tools  + Vision    & $\mathbf{0.27 \hspace{-0.5mm}\pm\hspace{-0.5mm} 0.12}$ & $\mathbf{0.89 \hspace{-0.5mm}\pm\hspace{-0.5mm} 0.09}$ & $\mathbf{0.20 \hspace{-0.5mm}\pm\hspace{-0.5mm} 0.15}$ \\
        \bottomrule
    \end{tabular}
    \vspace{-0.75\baselineskip}
\end{table*}

\section{Experiments and Results}

Our empirical validation systematically evaluates the core principles of our methodology. We conducted targeted experiments within SRB to deconstruct the complex problem of autonomous excavation. These experiments first establish the foundational role of adaptive compliance, then demonstrate the critical importance of procedural diversity for generalization, and finally quantify the benefits of visual perception.

\subsection{Experimental Setup}

Our investigation is structured as a direct comparison of distinct agents that differed in their training experience, compliance strategy, and available sensory observations. All experiments use a simulated Franka Emika robotic manipulator mounted on a stationary pedestal. The simulation is executed with a physics update rate of $250$~Hz, while the learning agent operates at a control frequency of $50$~Hz. To balance physical fidelity with computational cost, we use $16$ parallel environment instances, each populated with $100$k regolith particles. For each instance, a particle system is spawned in front of the robot in a pyramid formation above a unique procedural terrain. The initial velocity of each particle is sampled from a uniform distribution $U(-0.5, 0.5)$~m/s for the XY axes and $U(-0.5, 0.0)$~m/s for the Z axis. The particles are then allowed to settle under gravity for up to $300$~s or until their median velocity drops below a stable threshold of $0.01$~m/s. This settled state is then cached as the starting condition for all subsequent training episodes. During training, the physical properties of the regolith and the environment are varied through domain randomization. Each particle has a radius of $0.004$~m, a density sampled from $\mathcal{N}(1500, 200)$~kg/m$^3$, a friction coefficient from $\mathcal{N}(0.9, 0.1)$, and a cohesion value from $\mathcal{N}(0.1, 0.01)$. The global gravity magnitude is also sampled from $U(1.6123, 1.6376)$~m/s\textsuperscript{2} to simulate operations in different lunar regions. The robot begins each episode from a randomized joint configuration with its rigidly attached excavation tool positioned above the regolith pile.

All agents are trained using the DreamerV3 algorithm, which distinguishes between a large world model used during training and a more compact inference policy. The proprioceptive agents utilize an architecture with $86$M learnable parameters, with over $90$\% dedicated to the training-time world model, and the remaining $8$M are shared during inference by the encoder and the policy. The agent variant with visual observations incorporates additional convolutional layers to process depth map inputs, which results in a larger architecture with $130$M parameters ($10$M during inference). To achieve satisfactory sample efficiency with the computationally intensive particle simulation, we employ a large replay buffer with a capacity of $2.5$M transitions combined with a high training ratio of $256$. Other key hyperparameters are kept consistent with the original implementation to ensure a robust baseline~\cite{hafner2025mastering}. The proprioceptive agents were trained for a total of $5$M environment steps with a mean wall-clock time of $71$~h on a single NVIDIA RTX 4090 GPU. Due to the increased computational load of processing high-dimensional image data, the \textsc{visual} agent took $218$~h to complete training. The resulting inference latency for the final policies was $1.56 \hspace{-0.5mm}\pm\hspace{-0.5mm} 0.1$~ms for the proprioceptive agents and $2.08 \hspace{-0.5mm}\pm\hspace{-0.5mm} 0.2$~ms for the \textsc{visual} agent. While we acknowledge that the high-performance hardware acceleration is not yet space-qualified, this limitation is being addressed by rapid advancements in onboard computing~\cite{felix2024total}. Additionally, the policy could be further optimized using established techniques like quantization and pruning.

The final evaluation of all trained agents is conducted on a held-out test set of $8$ novel tool geometries that were not seen during training. As shown in Figure~\ref{fig:tool_comparison}, this set includes $4$ procedural variants and $4$ manual designs to ensure a comprehensive assessment of generalization. We measure performance using three key metrics. The first is the mean volume of successfully excavated regolith, quantified as the normalized count of lifted and stabilized particles at the end of an episode. The second metric is the amount of generated dust, computed as the cumulative fraction of particles exceeding a velocity threshold of $0.02$~m/s throughout the episode. The third metric is control smoothness, measured by the time-averaged mean squared jerk of the robot's joint motion. Each reported result is an average over $50$ independent evaluation episodes for each of the $8$ distinct tool geometries.

\subsection{Results and Discussion}

\begin{figure}[ht]
    \centering
    \includegraphics[width=\linewidth]{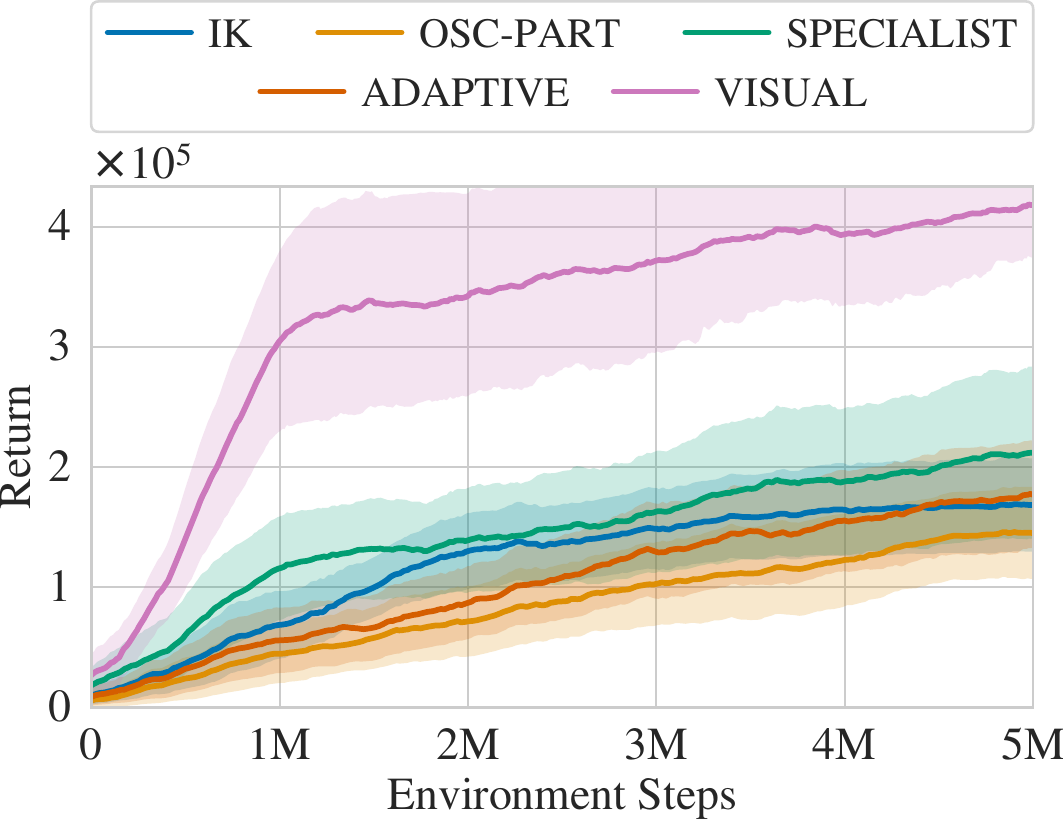}
    \caption{Learning curves of evaluated agents during training, averaged over three seeds (two for \textsc{visual}), with shaded regions representing the standard deviation.}
    \label{fig:main_learning_curves}
    \vspace{-0.75\baselineskip}
\end{figure}

Our experiments were designed to systematically deconstruct the challenge of autonomous excavation by isolating the key factors that contribute to a robust policy. First, we establish the role of adaptive compliance for managing complex physical interactions. We then demonstrate the critical role of procedural diversity in training for generalization. Finally, we quantify the performance benefits of augmenting the agent with visual perception and analyze the implications of our results for robotic tool design. The learning curves of all evaluated agents are presented in Figure~\ref{fig:main_learning_curves}, and their final performance on the held-out set of novel tools is summarized in Table~\ref{tab:main_results}.

\paragraph{Impact of Adaptive Compliance}
To manage the unpredictable forces of excavation, a robot must interact compliantly with its environment. Our first experiment confirms this necessity by comparing a rigid kinematic controller with agents that could learn to modulate their own compliance. As shown in Table~\ref{tab:main_results}, all three agents ultimately learned to excavate a similar volume of regolith. However, the critical difference is in the quality and safety of the interaction. While successful in excavating material, the rigid \textsc{ik} agent achieved its goal through aggressive and uncontrolled interactions. This resulted in significantly jerkier motions and the generation of more dust. In contrast, the \textsc{adaptive} agent, which learned to control both stiffness and damping, demonstrated a markedly smoother and more controlled interaction style. It produced the least amount of dust, which indicates a safer and more stable approach. The \textsc{osc-part} agent offered slightly smoother motion but was outperformed in the amount of excavated regolith. This result demonstrates that while rigid control can succeed, learned adaptive compliance is essential for performing the task safely and efficiently, which is a prerequisite for long-term hardware reliability in space.

\paragraph{Generalization through Procedural Diversity}
Having established the need for compliance, our central experiment investigated how to create a truly generalizable excavation skill. We compared a \textsc{specialist} agent trained with a single, static tool geometry against a generalist (\textsc{adaptive}) agent trained on the full procedural distribution of tools. The learning curves in Figure~\ref{fig:main_learning_curves} show that the \textsc{specialist} learned more quickly since it only needed to master one condition. The generalist required a longer training period to master the diverse tool geometries. However, the evaluation on the held-out set of novel tools, shown in Table~\ref{tab:main_results}, revealed a critical flaw of the \textsc{specialist}. Its performance collapsed when presented with any tool different from the one it was trained on. The generalist agent, in contrast, maintained a high level of performance across all unseen tools. For completeness, we also evaluated both agents on the specific tool geometry used during the \textsc{specialist}'s training. The \textsc{specialist} unsurprisingly maintained its performance from training with a mean excavated volume of $0.15 \hspace{-0.5mm}\pm\hspace{-0.5mm} 0.06$~L, while the generalist agent performed slightly worse but still competently with $0.12 \hspace{-0.5mm}\pm\hspace{-0.5mm} 0.04$~L. This outcome provides definitive evidence for our core hypothesis. Training with procedural diversity forces the agent to learn the underlying principles of excavation rather than memorizing a single tool-specific strategy, which is crucial for forging a truly robust and adaptable policy for the unpredictable conditions of space.

\begin{figure}[ht]
    \centering
    \includegraphics[width=\linewidth]{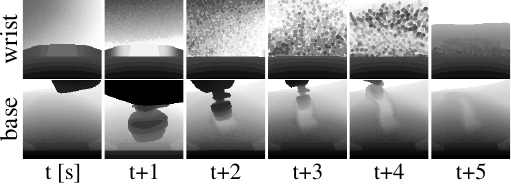}
    \caption{Visual feedback from two views over the course of a single excavation episode. Equalized for clarity.}
    \label{fig:vision_scenario}
    \vspace{-0.75\baselineskip}
\end{figure}

\paragraph{Effect of Visual Perception}
While the proprioceptive agent can infer some information from contact forces, it is effectively blind to the terrain before it. Our final study explores the benefit of augmenting the agent with vision. The visual feedback, shown in Figure~\ref{fig:vision_scenario}, provides the agent with a rich, real-time understanding of its workspace, including the shape of the regolith pile and the geometry of its own tool. This added situational awareness proves to be a significant advantage. The learning curves in Figure~\ref{fig:main_learning_curves} show that the \textsc{visual} agent learned faster and converged to a higher final reward. The quantitative results in Table~\ref{tab:main_results} confirm this superiority across all metrics. The agent with vision not only excavated a larger volume of material but also did so more safely with less generated dust. This confirms that direct perception enables more effective planning and reactive control, which is crucial for navigating the complex and dynamic environment of an excavation site.

\begin{figure}[t]
    \centering
    \includegraphics[width=\linewidth]{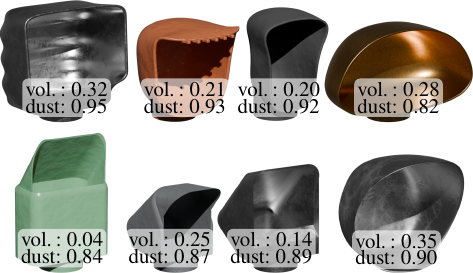}
    \caption{The performance of the \textsc{visual} agent across the held-out set of novel tool geometries [vol. unit: L].}
    \label{fig:tool_comparison}
    \vspace{-0.75\baselineskip}
\end{figure}

\paragraph{Implications for Tool Design}
Our methodology also presents a novel opportunity to inform hardware development. Predicting the most effective tool design is not straightforward, as each geometry presents a unique interaction challenge, and its performance is difficult to quantify because every tool may require a completely different control strategy. Because our learned agent is generalizable, it serves as a consistent evaluation metric for different mechanical designs. We leverage this capability by analyzing the performance of the \textsc{visual} agent across the held-out set of novel tools, with each tool evaluated over $50$ independent episodes. The results, presented in Figure~\ref{fig:tool_comparison}, reveal a significant variance in performance attributable to tool geometry alone. Some designs consistently enabled the agent to excavate more material with lower dust generation, indicating a superior suitability for the task with a learned controller. For instance, tools with deeper cavities at their leading edge outperformed their shallower counterparts, while rounded designs generally led to less dust generation. Although this analysis is preliminary and hinges on the specific learned policy, it nonetheless highlights the potential of using a learned agent as a tool for evaluating and optimizing mechanical designs. This data-driven approach could enable a synergistic co-design process where hardware and control policies are developed in parallel to create tools that are inherently more compatible with autonomous systems.

The results from our experiments provide strong and cohesive evidence for the effectiveness of our integrated methodology. The primary finding is that a generalizable excavation skill is forged from the powerful synergy of procedural diversity and learned adaptive compliance. By training with a wide distribution of tools, the agents discovered the underlying physical principles of excavation. They then expressed this knowledge by dynamically modulating their physical interaction style to suit the specific tool at hand. Furthermore, this work also suggests a new paradigm for co-designing robotic hardware and software. By systematically analyzing which tool geometries lead to superior performance with a generalist agent, we can identify design features that are inherently more compatible with autonomous control. This creates a powerful opportunity to optimize hardware and policies in a unified, data-driven loop. These findings are promising and grounded in simulation, highlighting clear directions for future research. The critical next step is a rigorous sim-to-real validation on a physical robotic testbed. Furthermore, the \textsc{visual} agent faces the additional challenge of the perceptual sim-to-real gap, which remains a significant hurdle in robotics. Finally, the high computational cost of particle physics is a current limitation. Future research could explore more efficient simulation techniques to enable even larger and more complex training scenarios.

\section{Conclusion}

The successful establishment of a sustainable off-world presence requires autonomous systems capable of foundational ISRU tasks like regolith excavation. This paper presented an integrated solution to address the complex physics and hardware variability inherent to this task. This solution is built on the principle that robust autonomy is forged through diversity. By training a model-based RL agent within a procedurally diverse simulation of both terrains and tools, we have shown it is possible to learn a truly generalizable excavation skill. The agent masters this challenge by learning to modulate its own physical compliance dynamically. This mechanism allows it to develop sophisticated, tool-aware strategies that are robust to novelty and safer in execution. The findings presented here offer more than a successful policy. They provide a validated methodology for overcoming the uncertainties of extraterrestrial environments. They also provide a novel framework for the data-driven co-design of robotic tools and their controllers. While sim-to-real transfer remains a critical next step, this research contributes the foundational knowledge necessary to advance the operational readiness of the autonomous systems that will build our multiplanetary future.

\end{document}